\documentclass[10pt,twocolumn,letterpaper]{article}

\usepackage{cvpr}              %

\usepackage[table,xcdraw]{xcolor}
\usepackage{graphicx}
\usepackage{svg}

\definecolor{cvprblue}{rgb}{0.21,0.49,0.74}

\usepackage{array}
\usepackage{times}
\usepackage{epsfig}
\usepackage{graphicx}
\usepackage{float}
\usepackage{wrapfig}
\usepackage{amsmath,amssymb,amsthm}
\usepackage{algorithm,algorithmicx,algpseudocode}
\usepackage{bm,xspace}
\usepackage{comment}
\usepackage{multirow}
\usepackage{balance}
\usepackage{url}
\usepackage{booktabs}
\usepackage{etoolbox,siunitx}
\usepackage{calc}
\usepackage{pifont,hologo}
\usepackage{color}
\usepackage{adjustbox}
\usepackage{enumitem}
\usepackage[normalem]{ulem}  %
\usepackage[table]{xcolor}
\usepackage[pagebackref,breaklinks,colorlinks,linkcolor=red,citecolor=blue]{hyperref}
\usepackage{natbib}
\PassOptionsToPackage{square,numbers,comma,sort}{natbib}

\newcommand{\authorskip}{\hspace{3mm}}
\newcommand{\institutionskip}{\hspace{6mm}}

\definecolor{blue}{HTML}{0055cc}
\definecolor{red}{HTML}{cc1100}
\definecolor{orange}{HTML}{cc7700}
\definecolor{gray}{HTML}{efefef}
\definecolor{darkgreen}{rgb}{0.13, 0.55, 0.13}
\definecolor{darkgray}{HTML}{757575}

\newcommand{\figref}[1]{Figure~\ref{#1}}
\newcommand{\tabref}[1]{Table~\ref{#1}}
\newcommand{\secref}[1]{Section~\ref{#1}}
\renewcommand{\eqref}[1]{Eq.~\ref{#1}}

\newcolumntype{x}[1]{>{\centering\arraybackslash}p{#1}}
\newcolumntype{y}[1]{>{\raggedright\arraybackslash}p{#1}}
\newcolumntype{z}[1]{>{\raggedleft\arraybackslash}p{#1}}
\newcommand{\tablestyle}[2]{\setlength{\tabcolsep}{#1}\renewcommand{\arraystretch}{#2}\centering\footnotesize}

\DeclareMathSymbol{@}{\mathord}{letters}{"3B}

\newcommand\mypara[1]{\vspace{0mm}\noindent\textbf{#1}}

\makeatletter
\DeclareRobustCommand\onedot{\futurelet\@let@token\@onedot}
\def\@onedot{\ifx\@let@token.\else.\null\fi\xspace}

\newcommand*{\Rom}[1]{\expandafter\@slowromancap\romannumeral #1@}
\newcommand*{\rom}[1]{\expandafter\romannumeral #1}

\def\1{\bm{1}}

\DeclareMathAlphabet{\mathsfit}{\encodingdefault}{\sfdefault}{m}{sl}
\SetMathAlphabet{\mathsfit}{bold}{\encodingdefault}{\sfdefault}{bx}{n}

\newcommand{\softmax}{\mathrm{softmax}}

\def\paperID{9400} %
\def\confName{CVPR}
\def\confYear{2024}

\title{GroupContrast: Semantic-aware Self-supervised Representation Learning \\ for 3D Understanding}

\author{Chengyao Wang\textsuperscript{\mdseries1} \authorskip 
Li Jiang\textsuperscript{3} \authorskip
Xiaoyang Wu\textsuperscript{\mdseries2} \authorskip 
Zhuotao Tian\textsuperscript{\mdseries4} \\
Bohao Peng\textsuperscript{1} \authorskip
Hengshuang Zhao\textsuperscript{2*} \authorskip
Jiaya Jia\textsuperscript{1,4} 
 \\ \\
\textsuperscript{1}CUHK \institutionskip
\textsuperscript{2}HKU \institutionskip
\textsuperscript{3}CUHK(SZ) \institutionskip
\textsuperscript{4}SmartMore \\
{\tt\small \url{https://github.com/dvlab-research/GroupContrast}}
}

\begin{document}
\maketitle
\renewcommand{\thefootnote}{\fnsymbol{footnote}}
\footnotetext[1]{Corresponding author.}
\begin{abstract}
Self-supervised 3D representation learning aims to learn effective representations from large-scale unlabeled point clouds.
Most existing approaches adopt point discrimination as the pretext task, which assigns matched points in two distinct views as positive pairs and unmatched points as negative pairs.
However, this approach often results in semantically identical points having dissimilar representations, leading to a high number of false negatives and introducing a ``\textit{semantic conflict}'' problem.
To address this issue, we propose GroupContrast, a novel approach that combines segment grouping and semantic-aware contrastive learning.
Segment grouping partitions points into semantically meaningful regions, which enhances semantic coherence and provides semantic guidance for the subsequent contrastive representation learning. 
Semantic-aware contrastive learning augments the semantic information extracted from segment grouping and helps to alleviate the issue of ``\textit{semantic conflict}''.
We conducted extensive experiments on multiple 3D scene understanding tasks. The results demonstrate that GroupContrast learns semantically meaningful representations and achieves promising transfer learning performance.
\end{abstract}
    
\section{Introduction}
\label{sec:intro}
Self-supervised visual representation learning aims to learn effective representations from large-scale unlabeled data. The learned representation can boost large amounts of downstream applications, such as object detection and semantic segmentation. Despite self-supervised learning has achieved remarkable results in 2D dense prediction tasks, 3D representation learning remains an emerging field. The predominant approaches for 3D scene recognition rely on supervised learning, where models are trained from scratch on specific datasets and tasks.

\begin{figure}[t]\centering
\includegraphics[width=1.0\linewidth]{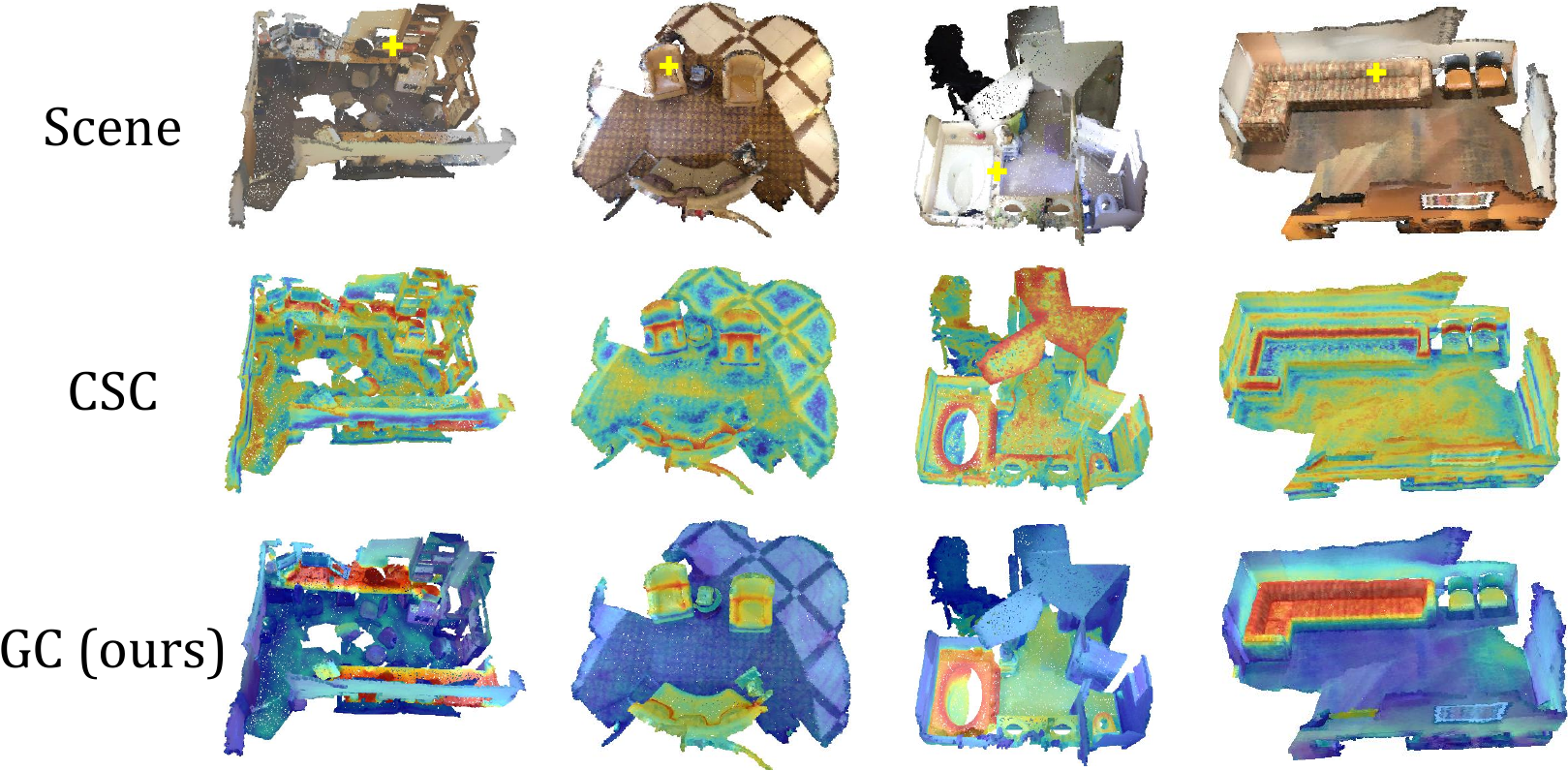}
\caption{
Visualization of activation maps depicting cosine similarity to the query point (indicated by a yellow cross) in the scene. Our approach demonstrates superior effectiveness in discriminating semantically similar points compared to CSC~\cite{csc}.
}
\label{fig:heatmap}
\end{figure}

Recent studies~\cite{pointcontrast, csc, msc} have explored the application of point discrimination as a pretext task for 3D self-supervised representation learning. 
These approaches transform each scene into two distinct views. They consider matched points between the two views as positive pairs and unmatched points as negative pairs. 
Despite decent performance gains observed in downstream tasks fine-tuning, there is a significant issue in previous point discrimination works: they focus on the geometric and structural relationships, disregarding the inherent semantic correlations. Consequently, they struggle to generate similar representations for semantically similar points within the 3D scene.
To illustrate this issue, we visualize the activation map of a previous point discrimination model, CSC~\cite{csc}, as shown in~\figref{fig:heatmap}.
The visualization reveals that previous methods fail to effectively capture semantic similarity. 
High-response points are scattered throughout the scene despite not necessarily possessing semantic similarity with query points. Conversely, points that are semantically similar to the query point may exhibit low correlations.

This observation motivates us to improve the point discrimination pretext task. Merely designating unmatched points as negative pairs in the pretext task may result in a high number of false negatives. This is because elements that should be semantically identical are compelled to have dissimilar representations, which we refer to as the ``\textit{semantic conflict}''.
As a consequence, the conflict may compromise the semantic consistency that is crucial for downstream dense prediction tasks where different individuals are required to be assigned to their corresponding semantic labels.
To address this issue, we present GroupContrast which consists of two essential parts: 1) Segment Grouping and 2) Semantic-Aware Contrastive Learning.

\vspace{2mm}
\mypara{Segment grouping} aims to enhance the semantic coherence among points within a scene and provide semantic guidance for the following contrastive learning. It achieves this by partitioning the point cloud into semantically similar groups via a segment-level deep clustering process. 
In particular, we first generate initial segments via a graph cut method based on low-level geometric information~\cite{graph_segment} and get the segment features via segment-wise pooling.
We employ a set of learnable prototypes as cluster centers.
Correlations between segment representations and these prototypes are then computed, and an informative-aware distillation loss is applied to encourage consistency between the segment-prototype correlations across two views with different augmentations.
Segment grouping holds significant potential for effectively grouping semantically similar segments, thereby serving as a robust foundation for advancing point discrimination and addressing ``\textit{semantic conflict}''.

\vspace{2mm}
\mypara{Semantic-aware contrastive learning.}
Based on the results of segment grouping, we can improve the pretext task of point discrimination by integrating the positive pairs obtained within the same group and the negative pairs derived from different groups. This approach helps to alleviate the issue of ``\textit{semantic conflict}'' by ensuring that the elements in negative pairs have distinct geometric representations in the representation space. An InfoNCE loss~\cite{cpc} is then applied to aggregate positive pairs and scatter negative pairs in the representation space. Besides, the confidence weight is found conducive to contrastive learning by mitigating the adverse impacts of incorrect segment assignments yielded by segment grouping.

As shown in the activation map depicted in~\figref{fig:heatmap}, our method effectively recognizes semantically similar points in the scene for the query point, in contrast to the confusion observed in the CSC model~\cite{csc}. This highlights the emerging capacity of GroupContrast in semantic-level recognition.
Extensive experiments on 3D semantic segmentation, instance segmentation, and object detection demonstrate the promising transfer learning performance of GroupContrast. 
For instance, our approach achieves 75.7\% mIoU on ScanNet~\cite{scannet} and 30.0\% mIoU on ScanNet200~\cite{scannet200} semantic segmentation using a SparseUNet~\cite{spareseunet} pre-trained by our method.
These results outperform current state-of-the-art self-supervised 3D representation learning approaches.
The contribution of our work can be summarized as follows:
\begin{itemize}[leftmargin=5mm, itemsep=0mm, topsep=-1mm, partopsep=-1mm]
    \item We examine the representations generated by the current unsupervised point cloud representation learning method and observe the presence of semantic conflict, which can potentially impede the performance of downstream applications.
    \item We propose GroupContrast, consisting of Segment Grouping and Semantic-aware Contrastive Learning, to address the semantic conflict by preserving the cross-view geometric consistency while avoiding negative pairs with similar semantics.
    \item Extensive experiments demonstrate that GroupContrast achieves state-of-the-art transfer learning results in various 3D scene perception tasks.
\end{itemize}

\section{Related Work}

\begin{figure*}[t]\centering
\includegraphics[width=1.0\linewidth]{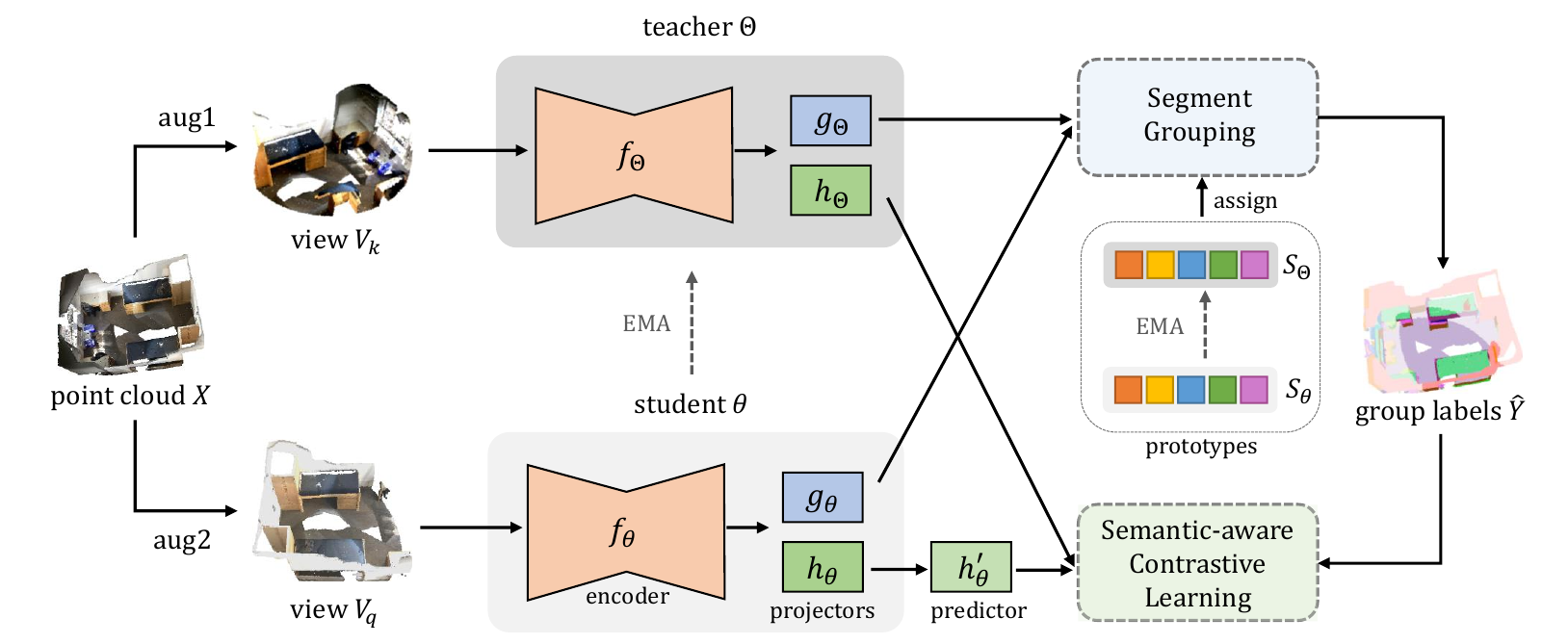}
\caption{\textbf{Overview of our proposed GroupContrast framework.}
Our framework uses two neural networks, each comprising a backbone and two projectors for segment grouping and contrastive learning. The parameters of the teacher network are updated as an exponential moving average (EMA) of the parameters of the student network. The student network includes an additional asymmetric predictor for contrastive learning. 
The Segment Grouping module assigns each point to one of $n$ prototypes, and this clustering result serves as a guide for effective contrastive representation learning.
}
\label{fig:framework}
\end{figure*}

\vspace{1mm}
\mypara{2D self-supervised representation learning.}
Instance discrimination~\cite{instance_discriminative} as a pretext task for self-supervised visual representation learning has made remarkable progress in recent years. 
By leveraging InfoNCE loss~\cite{cpc} as an optimization objective for contrastive representation learning, a number of studies~\cite{moco, mocov2, simclr} have shown impressive transfer learning performance. More recently, modern approaches~\cite{byol, swav, simsiam, dino, dinov2, mocov3} further improve this paradigm by removing negative pairs.
Despite the impressive transfer learning performance on image classification tasks, instance discrimination treats an image as a whole, migrating complex structures in natural images. To address this issue, some studies explore pixel discrimination~\cite{densecl, pixpro, self_embed} and object discrimination~\cite{detcon, orl, soco, slotcon, odin} as a pretext task, which enhances the intrinsic structure of the image and further improves the transfer learning performance on dense prediction tasks.
In this work, we attempt to conduct visual representation learning on complicated 3D scenes, which is more correlated with this line of work.

\vspace{1mm}
\mypara{3D self-supervise representation learning.}
Unlike the 2D counterpart, self-supervised representation learning on 3D point clouds is still an emerging area. Earlier works~\cite{wang2019deep, hassani2019unsupervised, sauder2019self, sanghi2020info3d} conduct self-supervised representation learning on object-centric point clouds~\cite{shapenet}. Experimentally, these approaches are unable to benefit 3D scene understanding~\cite{pointcontrast}. Recent works~\cite{pointcontrast, csc, depthcontrast, strl, msc, yang2024unipad, zhu2023ponderv2} start to build self-supervised 3D representation learning on scene-centric data~\cite{scannet} and found sufficient performance improvement when transferred to a diverse set of 3D scene perception tasks.
As a pioneer work, PointContrast~\cite{pointcontrast} adopts point discrimination as a pretext task. CSC~\cite{csc} explore point discrimination with scene context descriptors.
MSC~\cite{msc} introduces masked reconstruction learning to enforce the pretext and alleviate the mode collapse problem.
However, these approaches treat matched points as positive pairs and unmatched points as negative pairs, leading to a large number of false negatives. In contrast, we attempt to discover semantic meaningful regions to avoid the model being confused by false negatives.
Cluster3Dseg~\cite{feng2023clustering} group points with identical labels into subclass and learn a better representation space via contrastive learning. But they focus on supervised learning, while we focus on unsupervised representation learning.

\mypara{3D scene understanding.}
There are two primary architectures for 3D scene understanding: point-based and voxel-based methods.
Point-based methods~\cite{pointnet, pointnet++, pointweb, ptv1, ptv2, wu2024ptv3, yang2023sam3d, zhong2023understanding} directly operate on the points, making them well-suited for learning point clouds. However, directly operating on the points makes this line of work computationally expensive.
In contrast, voxel-based methods transform points into regular voxels to apply 3D convolutions~\cite{maturana2015voxnet,song2017semantic}. Driven by highly optimized sparse convolution~\cite{mink, spareseunet}, this line of work achieves excellent efficiency.
Following previous works on 3D representation learning~\cite{pointcontrast,msc,wu2024ppt}, we conduct representation learning and downstream task fine-tuning on a voxel-based method SparseUNet~\cite{spareseunet}.

\section{Method}

In this section, we introduce a novel method, GroupContrast, for 3D self-supervised representation learning to enhance the feature alignment among semantically similar points. GroupContrast consists of two key components: Segment Grouping and Semantic-aware Contrastive Learning. Firstly, we present the overall framework of our GroupContrast in~\secref{sec:framework}. Then, we delve into Segment Grouping in~\secref{sec:segment_grouping}, which enables the discovery of semantic meaningful regions in unlabeled point clouds. Following that, we introduce Semantic-aware Contrastive Learning in \secref{sec:contrastive_learning}, which leverages the regions discovered in Segment Grouping for effective contrastive representation learning.

\subsection{Overall Framework}
\label{sec:framework}

In GroupContrast, we employ a dual-network structure, including a teacher network and a student network, to ensure a stable and consistent contrastive learning process.
As illustrated in \figref{fig:framework}, the student network $\theta$ consists of an encoder $f_{\theta}$, two projectors  $g_{\theta}$ and $h_{\theta}$, an asymmetric predictor $h^{\prime}_{\theta}$, and a set of $n$ learnable prototypes $S_{\theta} \in \mathcal{R}^{n\times D}$, where $D$ indicates the feature dimension. 
The teacher network $\Theta$ shares the same architecture as the student network, except for the asymmetric prediction layer $h^{\prime}_{\theta}$. The teacher network has a different set of parameters, which are formed by taking the exponential moving average (EMA) of the student network $\theta$.

Given a point cloud $X$, two augmented views $V_k$ and $V_q$ derived from $X$ are fed into the teacher network and the student network, respectively. Then, we take the point-level features produced by the projectors $g$ and $h$ as the inputs for the Segment Grouping module and Semantic-aware Contrastive Learning module, respectively. 
In the Segment Grouping process, a set of learnable prototypes $S$ are used as cluster centers for identifying the meaningful semantic groups within the 3D scene. These groups are then employed in the Semantic-aware Contrastive Learning module to mitigate the semantic conflict problem and assist the representation learning.

\begin{figure}[t]\centering
\includegraphics[width=1.0\linewidth]{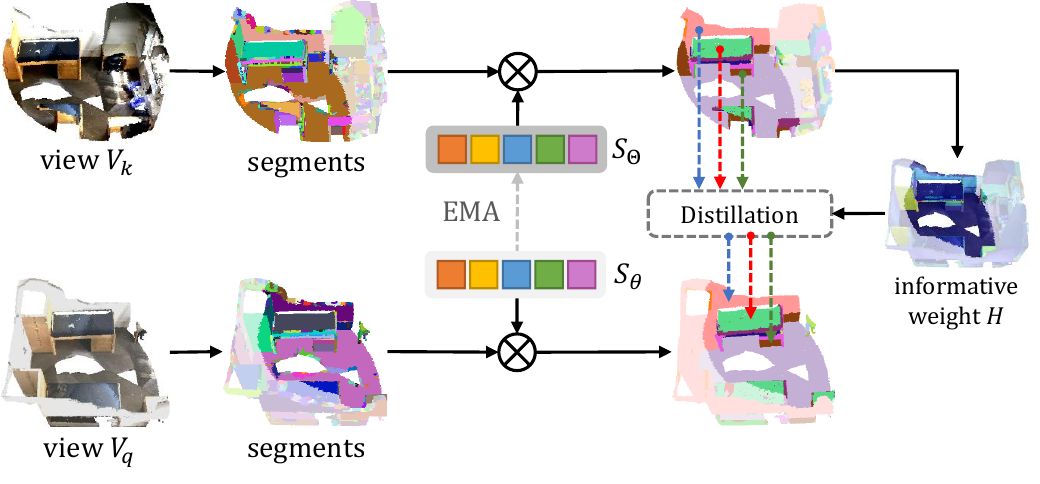}
\caption{\textbf{Segment Grouping} is optimized by distilling the assignment scores between each segment and the $n$ prototypes from the teacher network to the student network. An informative weight is employed to make the student network focus on more challenging segments. }
\label{fig:grouping}
\end{figure}

\subsection{Segment Grouping}
\label{sec:segment_grouping}

The Segment Grouping module is illustrated in \figref{fig:grouping}. We first utilize geometric information of the points (\textit{e.g.}, normal) to generate $P$ segments for the overlapped region of the two augmented views $V_q$ and $V_k$ using a graph cut method~\cite{graph_segment}. Then, segment-wise average pooling is applied on the l2-normalized point-level features produced by the projector $g$ for both augmented views,
resulting in the segment-level features $z_q \in \mathcal{R}^{P \times D}$ and $z_k \in \mathcal{R}^{P \times D}$.
After that, we calculate the prototype assignment scores for each segment by measuring the cosine similarity between the 
segment-level features and the $n$ learnable prototypes. Specifically, with segment-level features $z_q$, $z_k$ and $n$ learnable prototypes $S_{\theta}$, $S_{\Theta}$ which are all l2-normalized, 
the assignment scores $Q \in \mathcal{R}^{P \times n}$ and $K \in \mathcal{R}^{P \times n}$ for segments in each view can be written as 

{
\fontsize{9}{11}\selectfont
\begin{equation}
Q = \underset{n}{\softmax} (z_q S_{\theta}^\mathrm{T} / \tau_s), \quad
K = \underset{n}{\softmax} (z_k (S_{\Theta} - c)^\mathrm{T} / \tau_t).
\label{eq:assignment}
\end{equation}
}
Here, the temperature parameters $\tau_t$ and $\tau_s$ control the sharpness of the output distribution for the teacher network and student network, respectively. Additionally, a bias term $c$ is introduced to avoid collapse, which will be further discussed later. 

The teacher network is an average of consecutive student networks. Averaging model weights over training steps tends to produce a more accurate model~\cite{acceleration, mean_teacher}. We can take advantage of this to set the optimization objective of segment grouping. A cross-entropy loss is applied to encourage the assignment scores of the student network consistent with the teacher network.
The overall grouping loss is computed as the average cross-entropy loss across $P$ segments:
\begin{equation}
\mathcal{L}^{group} = - \frac{1}{P} \sum_{i \in [0, P)} K_i \cdot log(Q_i).
\label{eq:distill}
\end{equation}

\vspace{1mm}
\mypara{Prevention of collapse.} 
Directly applying this optimization objective will lead to collapse~\cite{dino}.
Inspired by DINO~\cite{dino}, we apply centring and sharpening for the momentum teacher outputs to avoid model collapse.
For sharpening, we make the teacher temperature $\tau_t$ lower than student temperature $\tau_s$ to produce a sharper target to avoid uniform assignments.
For centring, we use a bias term $c$ to the teacher and reduce it from the prototypes when producing the teacher assignments. The bias term $c$ is formed by taking the EMA of the output produced by the teacher network:
\begin{equation}
c = \lambda_c \cdot c + (1 - \lambda_c) \cdot \frac{1}{P} \sum_{i \in [0, P)} z_k\left[i\right] S_{\Theta}^\mathrm{T}, %
\label{eq:center}
\end{equation}
where $P$ stands for the number of segments,
and $\lambda_c$ refers to the momentum value. Intuitively, centring prevents one prototype from dominating the prototype assignment process.

\begin{figure}[t]\centering
\includegraphics[width=0.95\linewidth]{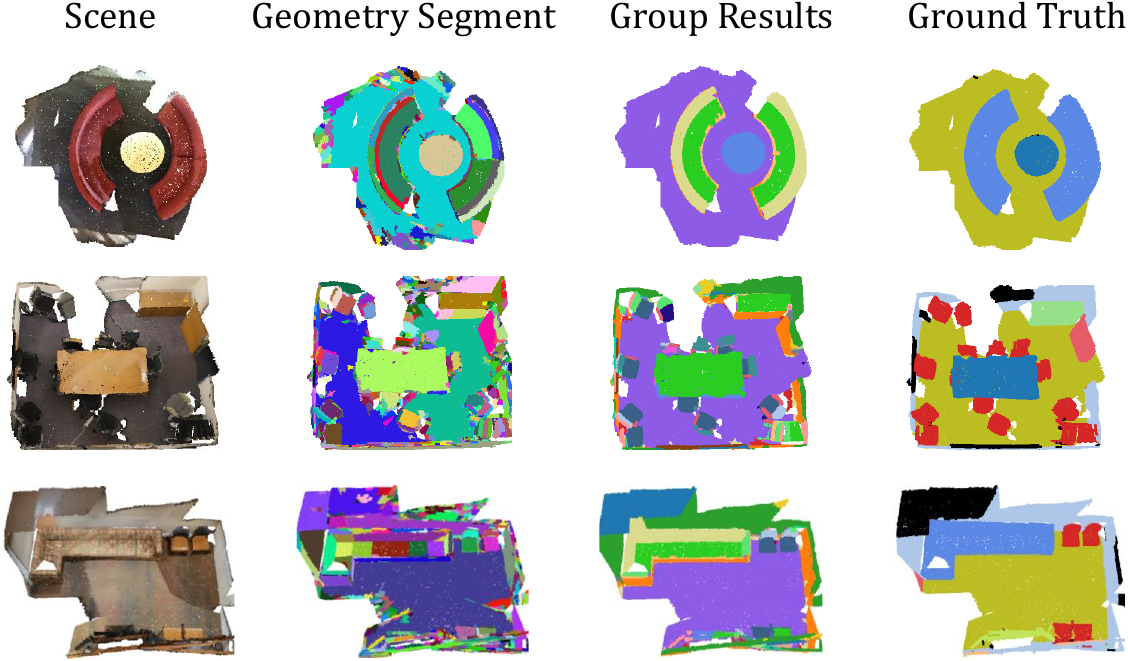}
\caption{\textbf{The result of Segment Grouping.} We compare the grouping results with original geometry segments~\cite{graph_segment} and semantic ground truth. Segment grouping effectively groups points into semantically meaningful regions without human supervision.}
\label{fig:group}
\end{figure}

\vspace{1mm}
\mypara{Informative-aware distillation.} The approach described above can be regarded as a knowledge distillation procedure from teacher network $\Theta$ to student network $\theta$. 
However, treating all segments equally in distillation can lead to the model overlooking more informative segments. These segments are typically more difficult to assign prototypes, \textit{i.e.}, with higher entropy, and should be paid with extra attention during distillation.
Therefore, we use the entropy of teacher assignment scores $K$ to measure each segment's ``amount of information'' and incorporate the entropy mask to distillation loss. The entropy mask $H$ for each segment $i$ can be concluded as $H_i = - \sum_{j=1}^n K_i^j \cdot log(K_i^j)$, and the grouping loss in \eqref{eq:distill} can be updated as
\begin{equation}
\mathcal{L}^{group} = - \frac{1}{\sum_{i \in [0, P)} H_i} \sum_{i \in [0, P)} H_i \cdot K_i \cdot log(Q_i).
\label{eq:distill2}
\end{equation}

\vspace{1mm}
\mypara{Grouping result.}
With the assignment scores, we group the $P$ segments into $n$ clusters by assigning each segment to the prototype with the highest assignment score. We use the teacher assignment scores $K$ to extract grouping results. Formally, the segment-level grouping result $\hat{Y}_{seg} \in \mathcal{R}^P$ is calculated as 
\begin{equation}
\hat{Y}_{seg} = \operatorname*{argmax}_{n} K.
\end{equation}
We then project $\hat{Y}_{seg}$ to each point to obtain the point-level group labels $\hat{Y}$ for the overlapped region of two views. As illustrated in~\figref{fig:group}, our segment grouping process effectively groups initial segments into semantically meaningful regions.

\begin{figure}[t]\centering
\includegraphics[width=0.75\linewidth]{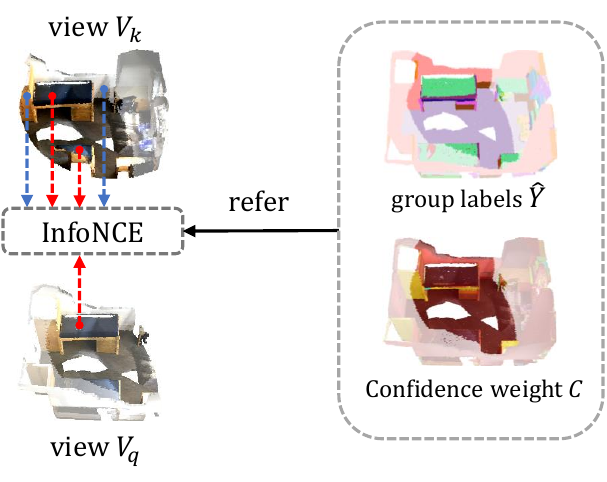}
\caption{\textbf{Contrastive Learning.} We use an InfoNCE loss~\cite{cpc} to aggregate points within the same group and scatter points across different groups, as indicated by the Segment Grouping result. Here the red point in view $V_q$ serves as a query, the red points in view $V_k$ are positive samples, and the blue points in view $V_k$ are negative samples. Both modules are conducted on overlapped regions of the two augmented views only, which are highlighted with darker colors in the figure.}
\label{fig:contrastive}
\end{figure}

\subsection{Semantic-aware Contrastive Learning}
\label{sec:contrastive_learning}
As discussed in Section~\ref{sec:intro}, the issue of ``\textit{semantic conflict}'' exists in previous contrastive-based representation learning methods where the semantically identical elements may erroneously have distinct representations.
To address this issue for achieving a better agreement between the semantically similar points, we use the group labels $\hat{Y}$ as the semantic guidance to enhance contrastive representation learning.

\vspace{1mm}
\mypara{Semantic-aware positive pairs.}
As illustrated in \figref{fig:contrastive}, we define the positive pairs for contrastive learning based on the semantic grouping result $\hat{Y}$. Points in the same group are set as positive pairs, while points in different groups are treated as negative pairs. Formally, for the two augmented views $V_q$ and $V_k$, we sample $N$ points from their overlapped region and set the point indices of these sampled points in $\hat{Y}$ as $I_q \in \mathcal{R}^N$ and $I_k \in \mathcal{R}^N$, respectively. The positive pair set is then defined as
\begin{equation}
    \mathcal{P} = \{(i, j) | i \in I_q, j \in I_k, \hat{Y}_i = \hat{Y}_j\}.
\end{equation}

\vspace{1mm}
\mypara{Confidence-aware learning.} In the early stages of training, the group labels $\hat{Y}$ may not always be reliable. Using noisy group labels for contrastive representation learning may confuse the model. Therefore, we evaluate the confidence of each positive pair and incorporate confidence weights in contrastive representation learning, to alleviate the adverse effects brought by the uncertain elements. 
Concretely, we leverage the teacher assignment scores $K$ in confidence evaluation. For each positive pair $i, j$ with grouping label $k$, the confidence weight $C_{i,j}$ is calculated as
\begin{equation}
C_{i,j} = K_{s_i, k} \times K_{s_j, k},
\end{equation}
where $s_i$ and $s_j$ indicate the segment indices of points $i$ and $j$, respectively.

\mypara{Improved contrastive loss.}
For the teacher network, we add a projector $h_{\Theta}$ after the encoder to extract feature $v_k$. For the student network, inspired by previous approaches~\cite{simclr, byol, dino}, a projector $h_{\theta}$ together with an extra asymmetric predictor $h^{\prime}_{\theta}$ is applied after encoder to extract feature $v_q$. Both $v_q$ and $v_k$ are l2-normalized for contrastive representation learning. 
The InfoNCE loss~\cite{cpc} is adopted to aggregate positive pairs and scatter negative pairs in the representation space. 
By incorporating the aforementioned confidence weight $C_{i, j}$, given a set of positive pairs $\mathcal{P}$ and a temperature parameter $\tau$, the improved contrastive loss can be written as

{
\fontsize{9}{11}\selectfont
\begin{equation}
\mathcal{L}^{con} = \frac{1}{|\mathcal{P}|}
\sum_{i,j \in \mathcal{P}}
{-} C_{i, j} \cdot \log \frac{\exp(v_q^i {\cdot} v_k^j / \tau)}{\exp(v_q^i {\cdot} v_k^j / \tau) {+} \sum\limits_{i,k \notin P}{\exp(v_q^i {\cdot} v_k^k / \tau)}}.
\label{eq:con}
\end{equation}
}
We set $\tau$ to 0.4, following previous approaches~\cite{pointcontrast, csc}.

\subsection{Overall Optimization Objective}
We jointly optimize segment grouping and contrastive representation learning for pre-training. The overall optimization objective is a weighted sum of \eqref{eq:distill2} and \eqref{eq:con}, which can be written as
\begin{equation}
\mathcal{L}^{overall} = \lambda_g \mathcal{L}^{group} + \lambda_c \mathcal{L}^{con},
\label{loss_pverall}
\end{equation}
where $\lambda_g$ and $\lambda_c$ are scale factors. 
We empirically set $\lambda_g=\lambda_c=1$, as our experiments suggest that the performance is robust to different scale factors.

\section{Experiments}

We conduct extensive experiments to validate the effectiveness of our proposed GroupContrast framework. 
We first perform ablation studies in \secref{sec:ablation} to demonstrate the efficacy of each proposed component, 
then compare our approach with previous state-of-the-art self-supervised 3D representation learning approaches in \secref{sec:compare}. 

\subsection{Main Properties}
\label{sec:ablation}

\begin{table*}[t!]
    \centering
    \subfloat[
        \textbf{Positive Pairs.} Positive pairs constructed based on Segment Grouping work best in downstream transfer.
        \label{tab:ablation_pseudo}
    ]{
        \begin{minipage}{0.35\linewidth}{\begin{center}
            \tablestyle{15pt}{1.05}
            \begin{tabular}{c|c}
\toprule
Positive Pairs & FT mIoU(\%) \\ \midrule
Matched Points & 74.6 \\
Spatial Grid & 74.2 \\
Geometry Segment & 74.8 \\
\rowcolor[HTML]{EFEFEF} 
Segment Grouping & \textbf{75.7} \\ \bottomrule
\end{tabular}%

        \end{center}}\end{minipage}
    }
    \hspace{3mm}
    \subfloat[
        \textbf{Number of prototypes.} 32 prototypes work best for Segment Grouping.
        \label{tab:ablation_prototype}
    ]{
        \begin{minipage}{0.28\linewidth}{\begin{center}
            \tablestyle{13pt}{1.05}
            \begin{tabular}{c|c}
\toprule
Prototypes $n$ & FT mIoU(\%) \\ \midrule
16 & 75.2 \\
\rowcolor[HTML]{EFEFEF} 
32 & \textbf{75.7} \\
64 & 75.3 \\
128 & 74.8 \\ \bottomrule
\end{tabular}

        \end{center}}\end{minipage}
    }
    \hspace{3mm}
    \subfloat[
        \textbf{Number of sampled points.} Our approach is robust to number of sampled points.
        \label{tab:ablation_sampled_points}
    ]{
        \begin{minipage}{0.28\linewidth}{\begin{center}
            \tablestyle{11pt}{1.05}
            \begin{tabular}{c|c}
\toprule
Sample Points & FT mIoU(\%) \\ \midrule
1024 & 75.5 \\
\rowcolor[HTML]{EFEFEF} 
2048 & \textbf{75.7} \\
4096 & 75.5 \\
8192 & 75.4 \\ \bottomrule
\end{tabular}

        \end{center}}\end{minipage}
    }
    \\
    \centering
    \subfloat[
        \textbf{Informative-aware.} Incorporating informative weight for distillation boosts downstream performance.
        \label{tab:ablation_entropy}
    ]{
        \begin{minipage}{0.35\linewidth}{\begin{center}
            \tablestyle{16pt}{1.05}
            \begin{tabular}{c|c}
\toprule
Infomative-aware & FT mIoU(\%) \\ \midrule
 & 75.4 \\
 \rowcolor[HTML]{EFEFEF} 
\checkmark & \textbf{75.7} \\ \bottomrule
\end{tabular}%

        \end{center}}\end{minipage}
    }
    \hspace{3mm}
    \subfloat[
        \textbf{Teacher temperature.} A softer teacher leads to better downstream performance.
        \label{tab:ablation_temp}
    ]{
        \begin{minipage}{0.28\linewidth}{\begin{center}
            \tablestyle{12pt}{1.05}
            \begin{tabular}{c|c}
\toprule
Temperature $\tau_t$ & FT mIoU(\%) \\ \midrule
0.04 & 75.3 \\
\rowcolor[HTML]{EFEFEF} 
0.07 & \textbf{75.7} \\ \bottomrule
\end{tabular}%

        \end{center}}\end{minipage}
    }
    \hspace{3mm}
    \subfloat[
        \textbf{Predictor.} Including asymmetric predictor boosts downstream performance.
        \label{tab:ablation_pred}
    ]{
        \begin{minipage}{0.27\linewidth}{\begin{center}
            \tablestyle{15pt}{1.05}
            \begin{tabular}{c|c}
\toprule
Predictor & FT mIoU(\%) \\ \midrule
 & 75.3 \\
 \rowcolor[HTML]{EFEFEF} 
\checkmark & \textbf{75.7} \\ \bottomrule
\end{tabular}%

        \end{center}}\end{minipage}
    }
    \\
    \centering
    \subfloat[
        \textbf{Semantic-aware Contrastive Learning.} Incorporating semantic-aware positive pairs and confidence weights can improve downstream performance.
        \label{tab:ablation_contrastive}
    ]{
        \begin{minipage}{0.35\linewidth}{\begin{center}
            \tablestyle{3pt}{1.2}
            \begin{tabular}{cc|c}
\toprule
Semantic-aware & Confidence-aware & FT mIoU(\%) \\ \midrule
 &  & 74.8 \\
\checkmark &  & 75.1 \\
\rowcolor[HTML]{EFEFEF} 
\checkmark & \checkmark & \textbf{75.7} \\ \bottomrule
\end{tabular}

        \end{center}}\end{minipage}
    }
    \hspace{3mm}
    \subfloat[
        \textbf{Avoid Collapse.} Incorporating both centering and sharpening helps our approach to alleviating the collapse problem.
        \label{tab:ablation_collapse}
    ]{
        \begin{minipage}{0.28\linewidth}{\begin{center}
            \tablestyle{5pt}{0.95}
            \begin{tabular}{cc|c}
\toprule
Centering & Sharpening & FT mIoU(\%) \\ \midrule
 &  & 72.7 \\
\checkmark &  & 73.1 \\
 & \checkmark & 74.1 \\
\rowcolor[HTML]{EFEFEF}
\checkmark & \checkmark & 75.7 \\ \bottomrule
\end{tabular}%

        \end{center}}\end{minipage}
    }
    \hspace{3mm}
    \subfloat[
        \textbf{Pre-training epochs.} Scaling up the number of pre-train epochs makes the model more data-efficient.
        \label{tab:ablation_epoch}
    ]{
        \begin{minipage}{0.27\linewidth}{\begin{center}
            \tablestyle{6pt}{1.2}
            \begin{tabular}{c|cc}
\toprule
Epochs & ScanNet & ScanNet~(20\%) \\ \midrule
300 & 74.8 & 65.0 \\
\rowcolor[HTML]{EFEFEF}
600 & \textbf{75.7} & 65.8 \\
1200 & \textbf{75.7} & \textbf{66.5} \\ \bottomrule
\end{tabular}%

        \end{center}}\end{minipage}
    }
    \caption{\textbf{Ablation Study.} Without further explanation, we pre-train a SparseUNet~\cite{spareseunet} for 600 epochs on ScanNet~\cite{scannet} dataset to analyse our main design choices and properties. We report fine-tuning mIoU results(\%) on ScanNet Semantic Segmentation. Default settings are marked in \colorbox{gray}{gray}.}
    \label{tab:ablation_study}
\end{table*}

To assess the effectiveness and analyze the key properties of our GroupContrast, we conduct ablation experiments on its core design choices. As a default setting, we first self-supervised pre-train a SparseUNet~\cite{spareseunet} on ScanNet~\cite{scannet} dataset for 600 epochs.
We then utilize ScanNet semantic segmentation as the downstream task and evaluate the performance using the mIoU (\%) metric. 
The results of ablation studies are concluded in \tabref{tab:ablation_study}.
Please refer to the supplementary material for more implementation details about the pre-training and fine-tuning.

\vspace{1mm}
\mypara{Positive pair construction.}
In \tabref{tab:ablation_pseudo}, we compare the semantic-aware positive pairs generated based on Segment Grouping results with several baselines to validate the effectiveness of Segment Grouping on contrastive representation learning. 
The baseline approaches include
(1) Matched Points, which uses matched points in two views as positive pairs and unmatched points as negative pairs, similar to PointContrast\cite{pointcontrast};
(2) Spatial Grid, which divides the point cloud into multiple spatial grids and assigns points within the same grid as positive pairs. In this case, we set the grid size to $1m \times 1m$; and
(3) Geometry Segment, which generates segments using a normal-based graph cut method~\cite{graph_segment} and assigns points within the same segment as positive pairs.
As shown, simply assigning points in the same spatial grid as positive pairs leads to a decrease in transfer learning performance, even worse than the Matched Points baseline. 
Although Geometry Segments incorporate geometric priors into the network, the improvement in transfer learning is only marginal. 
By employing Segment Grouping based on Geometry Segments, we observe a noteworthy improvement of 1.1 points in transfer learning performance compared to the Matched Points baseline, affirming the effectiveness of our proposed segment grouping approach.

\vspace{1mm}
\mypara{Number of prototypes.}
In \tabref{tab:ablation_prototype}, we investigate the impact of the number of prototypes $n$ in Segment Grouping. Our observations indicate that $n=32$ yields the best performance for ScanNet pre-training. Too few prototypes can lead to excessive feature aggregation, while too many can cause the model to learn overly fine-grained features.

\vspace{1mm}
\mypara{Number of sampled points.}
In \tabref{tab:ablation_sampled_points}, we study the number of points sampled from the overlapped region for contrastive representation learning. We observe that the model is robust to the number of sampled points. For training efficiency, we sample 2048 points for each 3D scene.

\vspace{1mm}
\mypara{Informative-aware distillation.}
In \tabref{tab:ablation_entropy}, we study the effect of informative-aware distillation. Introducing informative weight prevents the model from overlooking more informative segments during distillation, leading to better transfer learning performance.

\vspace{1mm}
\mypara{Teacher temperature $\tau_t$.}
In \tabref{tab:ablation_temp}, we study the temperature parameter in Segment Grouping. We follow previous work~\cite{dino} to set student temperature $\tau_s=0.1$ and ablate different values of teacher temperature $\tau_t$. 
The results show that a softer teacher leads to better transfer learning performance.

\vspace{1mm}
\mypara{Predictor.}
In \tabref{tab:ablation_pred}, we study the effect of the asymmetric predictor. Similar to previous 2D self-supervised representation learning approaches~\cite{simclr, byol, dino, slotcon}, introducing an asymmetric predictor makes the contrastive objective more challenging, resulting in better transfer learning performance.

\vspace{1mm}
\mypara{Semantic-aware contrastive representation learning.}
In \tabref{tab:ablation_contrastive}, we study the design of semantic-aware contrastive representation learning. Introducing semantic-aware positive pairs for contrastive learning helps alleviate the issue of ``\textit{semantic conflict}'' and results in better transfer learning performance. Moreover, incorporating confidence weight for each positive pair alleviates the adverse effects brought by the uncertain elements, further improving the transfer learning performance.

\begin{figure}[t]\centering
\includegraphics[width=1.0\linewidth]{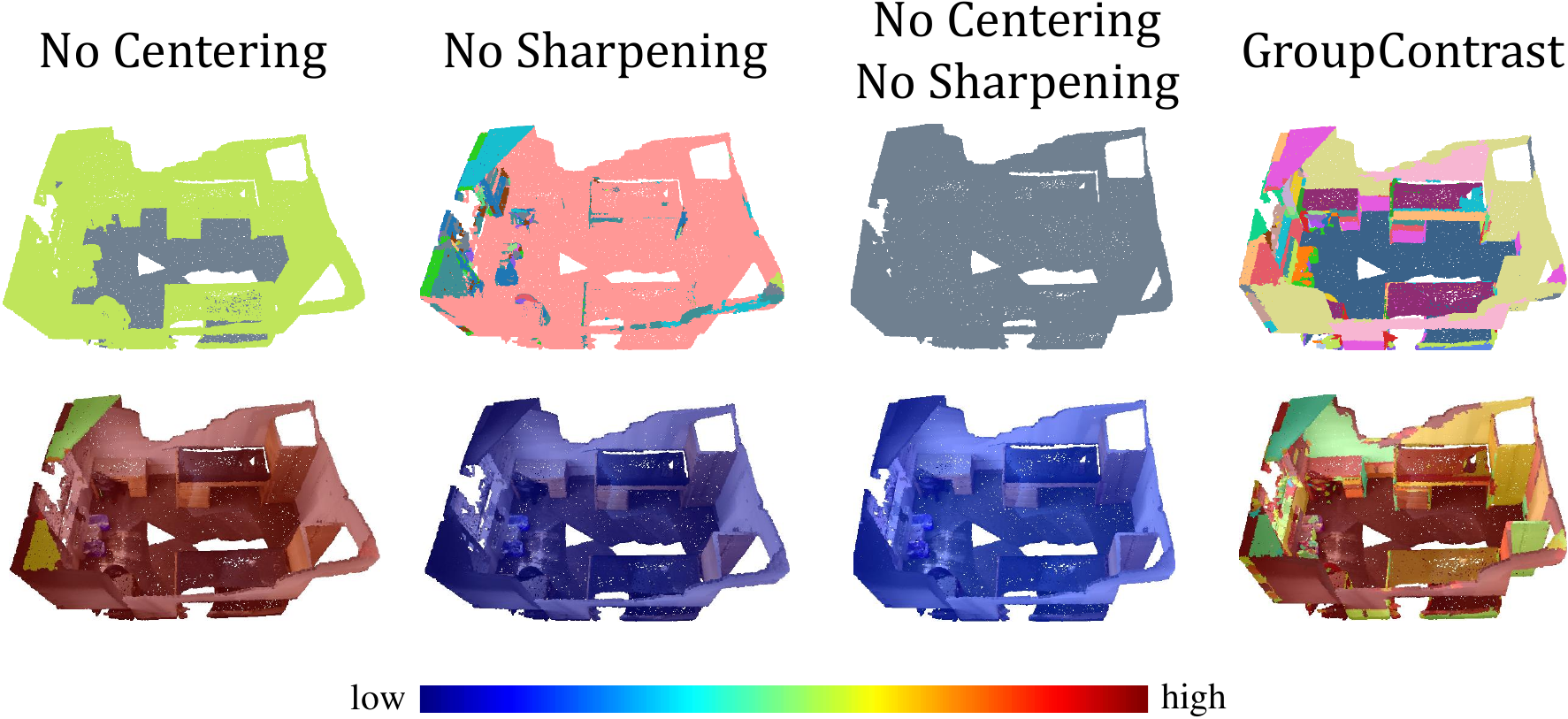}
\caption{\textbf{Group labels} (top) and \textbf{Confidence weight} (down) generated from the pre-trained model without centring and/or sharpening. Without centring, points are grouped into one or two regions. Without sharpening, the assignment scores become uniform vectors, leading to low confidence weight.}
\label{fig:collapse}
\end{figure}

\vspace{1mm}
\mypara{Collapse problem.}
In \tabref{tab:ablation_collapse}, we study the effectiveness of centring and sharpening in avoiding collapse. As shown in the table, both centring and sharpening effectively improve the transfer learning performance on semantic segmentation.
Furthermore, to further verify the effectiveness of centering and sharpening, we visualize the grouping results and corresponding assignment scores for GroupContrast without sharpening and centring in \figref{fig:collapse}.
Without centring, all points in a scene are grouped into two regions, resulting in multiple false positives for contrastive representation learning.
Without sharpening, the assignment scores $Q$ and $K$ become uniform vectors, resulting in low confidence scores and noisy group labels, which make it hard for contrastive loss to converge.
The result is even worse without both techniques, where all points are grouped into an identical region, and the corresponding confidence scores for each point are also low.

\vspace{1mm}
\mypara{Pre-training epochs.}
In \tabref{tab:ablation_epoch}, we study the number of pre-training epochs. Results indicate that increasing the pre-training epoch from 600 to 1200 does not enhance performance when fine-tuning the full ScanNet training set for semantic segmentation. However, in a data-efficient scenario where only 20\% of reconstructed point clouds are used for fine-tuning, scaling up the pre-training epoch effectively improves performance.

\begin{table}[t!]
    \centering
    \subfloat[
        \textbf{Semantic Segmentation.} We report the mIoU (\%) results on ScanNet, ScanNet200, and S3DIS benchmarks.
        \label{tab:exp_semseg}
    ]{
        \begin{minipage}{1.0\linewidth}{
        \begin{center}
            \tablestyle{6pt}{1.05}
            \begin{tabular}{c|ccccc}
\toprule
\multirow{2}{*}{Datasets} & \multicolumn{5}{c}{Semantic Segmentation (mIoU)} \\ \cmidrule{2-6} 
 & SC & PC\cite{pointcontrast} & CSC\cite{csc} & MSC\cite{msc} & \cellcolor[HTML]{EFEFEF}GC(ours) \\ \midrule
ScanNet & 72.2 & 74.1 & 73.8 & 75.3 & \textbf{\cellcolor[HTML]{EFEFEF}75.7} \\
ScanNet200 & 25.0 & 26.2 &26.4 & 28.8 & \textbf{\cellcolor[HTML]{EFEFEF}30.0} \\
S3DIS & 68.2 & 70.3 & \textbf{72.2} & - & \cellcolor[HTML]{EFEFEF}72.0 \\
\bottomrule
\end{tabular}%

        \end{center}
        }\end{minipage}
    }
    \\
    \subfloat[
        \textbf{Instance Segmentation.} We report the mAP@0.5 results on ScanNet, ScanNet200, and S3DIS benchmarks.
        \label{tab:exp_insseg}
    ]{
        \begin{minipage}{1.0\linewidth}{
        \begin{center}
            \tablestyle{6pt}{1.05}
            \begin{tabular}{c|ccccc}
\toprule
\multirow{2}{*}{Datasets} & \multicolumn{5}{c}{Instance Segmentation (mAP@0.5)} \\ \cmidrule{2-6} 
 & SC & PC\cite{pointcontrast} & CSC\cite{csc} & MSC\cite{msc} & \cellcolor[HTML]{EFEFEF}GC(ours) \\ \midrule
ScanNet & 56.9 & 58.0 & 59.4 & 59.6 & \textbf{\cellcolor[HTML]{EFEFEF}62.3} \\
ScanNet200 & 24.5 & 24.9 &25.2 & 26.8 & \textbf{\cellcolor[HTML]{EFEFEF}27.5} \\
S3DIS & 59.3 & 60.5 & 63.4 & - & \textbf{\cellcolor[HTML]{EFEFEF}63.5} \\
\bottomrule
\end{tabular}%

        \end{center}
        }\end{minipage}
    }
    \\
    \subfloat[
        \textbf{Object Detection.} We report the mAP@0.5 results on ScanNet and SUN RGB-D benchmarks.
        \label{tab:exp_detection}
    ]{
        \begin{minipage}{1.0\linewidth}{
        \begin{center}
            \tablestyle{6pt}{1.05}
            \begin{tabular}{c|ccccc}
\toprule
\multirow{2}{*}{Datasets} & \multicolumn{5}{c}{Object Detection (mAP@0.5)} \\ \cmidrule{2-6} 
 & SC & PC\cite{pointcontrast} & CSC\cite{csc} & MSC\cite{msc} & \cellcolor[HTML]{EFEFEF}GC(ours) \\ \midrule
ScanNet & 35.2 & 38.0 & 39.3 & - & \textbf{\cellcolor[HTML]{EFEFEF}41.1} \\
SUN RGB-D & 31.7 & 34.8 & 36.4 & - & \textbf{\cellcolor[HTML]{EFEFEF}37.0} \\
\bottomrule
\end{tabular}%

        \end{center}
        }\end{minipage}
    }
    \caption{\textbf{Results comparison} on 3D Semantic Segmentation, Instance Segmentation and Object Detection. We pre-train our approach on ScanNet point cloud with SparseUNet\cite{spareseunet} as the backbone for transfer learning performance comparison. \textit{SC} denotes train from scratch. Our results are marked in \colorbox{gray}{gray}. }
    \label{tab:transfer}
\end{table}

\subsection{Results Comparison}
\label{sec:compare}

In this section, we evaluate the effectiveness of our proposed GroupContrast by comparing it with previous self-supervised 3D representation learning approaches~\cite{pointcontrast, csc, msc}.
Experiments are conducted on various downstream tasks, including 3D semantic segmentation, instance segmentation, and object detection.
Additionally, we evaluate the data efficiency of GroupContrast on data-efficient 3D semantic segmentation.
We apply SparseUNet~\cite{spareseunet} as the backbone and adopt a longer training schedule (1200 epochs).
The transfer learning results are concluded in \tabref{tab:transfer}. Please refer to supplementary materials for more implementation details about downstream task fine-tuning.

\vspace{1mm}
\mypara{Semantic segmentation.}
In \tabref{tab:exp_semseg}, we report the Semantic Segmentation results on ScanNet~\cite{scannet}, ScanNet200~\cite{scannet200} and S3DIS~\cite{s3dis} benchmark with a SparseUNet backbone. 
For in-domain transfer learning where pre-training and fine-tuning are both conducted on the ScanNet~\cite{scannet} dataset, we achieve 75.7\% mIoU on the ScanNet validation set and 30.0\% mIoU on the ScanNet200 validation set, outperforming the current state-of-the-art approaches by 0.7\% mIoU and 1.2\% mIoU, respectively. 
Moreover, the model pre-trained with our approach on ScanNet achieves 72.0\% mIoU when transferred to S3DIS semantic segmentation, demonstrating that GroupContrast is also effective for cross-domain transfer learning.

\begin{table}[t!]
    \centering
    \subfloat[
        \textbf{Limited Reconstruction.} We compare the mIoU (\%) results on ScanNet data efficient semantic segmentation benchmark with limited scene reconstruction setting.
        \label{tab:exp_efficient_lr}
    ]{
        \begin{minipage}{1.0\linewidth}{
        \begin{center}
            \tablestyle{11pt}{1.05}
            \begin{tabular}{c|cccc}
\toprule
LR & \multicolumn{4}{c}{Semantic Segmentation (mIoU)} \\ \midrule
Pct. & SC & CSC~\cite{csc} & MSC~\cite{msc} & \cellcolor[HTML]{EFEFEF}GC~(ours) \\ \midrule
100\% & 72.2 & 73.8 & 75.0 & \textbf{\cellcolor[HTML]{EFEFEF}75.7} \\
1\% & 26.1 & 28.9 & 29.2 & \textbf{\cellcolor[HTML]{EFEFEF}30.7} \\
5\% & 47.8 & 49.8 & 50.7 & \textbf{\cellcolor[HTML]{EFEFEF}52.9} \\
10\% & 56.7 & 59.4 & 61.0 & \textbf{\cellcolor[HTML]{EFEFEF}62.0} \\
20\% & 62.9 & 64.6 & 64.9 & \textbf{\cellcolor[HTML]{EFEFEF}66.5} \\ \bottomrule
\end{tabular}%

        \end{center}
        }\end{minipage}
    }
    \\
    \subfloat[
        \textbf{Limited Annotation.} We compare the mIoU (\%) results on ScanNet data efficient semantic segmentation benchmark with limited point annotation setting.
        \label{tab:exp_efficient_la}
    ]{
        \begin{minipage}{1.0\linewidth}{
        \begin{center}
            \tablestyle{11pt}{1.05}
            \begin{tabular}{c|cccc}
\toprule
LA & \multicolumn{4}{c}{Semantic Segmentation (mIoU)} \\ \midrule
Pct. & SC & CSC~\cite{csc} & MSC~\cite{msc} & \cellcolor[HTML]{EFEFEF}GC~(ours) \\ \midrule
Full & 72.2 & 73.8 & 75.0 & \textbf{\cellcolor[HTML]{EFEFEF}75.7} \\
20 & 41.9 & 55.5 & \textbf{61.2} & \textbf{\cellcolor[HTML]{EFEFEF}61.2} \\
50 & 53.9 & 60.5 & 66.8 & \textbf{\cellcolor[HTML]{EFEFEF}67.3} \\
100 & 62.2 & 65.9 & 69.7 & \textbf{\cellcolor[HTML]{EFEFEF}70.3} \\
200 & 65.5 & 68.2 & 70.7 & \textbf{\cellcolor[HTML]{EFEFEF}71.8} \\ \bottomrule
\end{tabular}%

        \end{center}
        }\end{minipage}
    }
    \caption{\textbf{Data Efficiency.} We evaluate the data efficiency of GroupContrast on ScanNet data efficient semantic segmentation benchmark. The model is pre-trained on ScanNet point cloud with SparseUNet\cite{spareseunet} as the backbone. \textit{SC} denotes train from scratch. Our results are marked in \colorbox{gray}{gray}. }
    \label{tab:efficiency}
\end{table}

\vspace{1mm}
\mypara{Instance segmentation.}
In \tabref{tab:exp_insseg}, we report the instance segmentation results on ScanNet~\cite{scannet}, ScanNet200~\cite{scannet200} and S3DIS~\cite{s3dis} benchmark with PointGroup~\cite{pointgroup} as the instance segmentation head and SparseUNet as the backbone. 
We still find consistent transfer learning performance improvement compared with previous results. Specifically, our approach achieves 62.3 mAP@0.5 on the ScanNet validation set, which is 2.7 points higher than the previous state-of-the-art 3D self-supervised pre-training method. Furthermore, our approach achieves 27.5 mAP@0.5 when fine-tuning on ScanNet200 instance segmentation and 63.5 mAP@0.5 when fine-tuning on S3DIS instance segmentation, both outperforming previous 3D self-supervised pre-training approaches.

\vspace{1mm}
\mypara{Object detection.}
In \tabref{tab:exp_detection}, we report the Object Detection results on ScanNet~\cite{scannet} and SUN-RGBD~\cite{sun_rgbd} benchmark with VoteNet~\cite{votenet} as the detection head and SparseUNet as the backbone.
As shown in the table, we can also find performance improvement in object detection fine-tuning. We can achieve 41.1 mAP@0.5 on the ScanNet validation set and 37.0 mAP@0.5 on the SUN-RGBD validation set, surpassing the previous state-of-the-art 3D self-supervised pre-training approach by 1.8 points and 0.6 points respectively.

\vspace{1mm}
\mypara{Data efficiency.}
Apart from full dataset fine-tuning, we also evaluate the data efficiency of our approach on the ScanNet Data Efficient Semantic Segmentation benchmark. We use the same data split as CSC~\cite{csc} in both limited scene reconstruction and limited points annotation settings.
We report the data efficient semantic segmentation result in \tabref{tab:efficiency}. As reported, our approach achieves state-of-the-art performance for all cases in both settings.
These results suggest that our proposed approach is effective in improving the data efficiency of 3D scene understanding.

\section{Conclusion}
This work presents GroupContrast, a self-supervised representation learning framework for 3D scene understanding, with joint segment grouping and semantic-aware contrastive learning. 
Segment grouping discovers semantically meaningful regions by assigning each segment a prototype. Based on the grouping result, a contrastive learning objective is then applied to produce a semantic-aware representation space.
Our approach can effectively decompose a point cloud into multiple semantically meaningful regions without supervision, showing the emerging ability in semantic-level recognition.
Moreover, extensive experimental results demonstrate that our approach achieves promising transfer learning performance on various 3D scene understanding tasks, such as 3D semantic segmentation, object detection and instance segmentation. 

While our GroupContrast brings great benefits to downstream tasks through ScanNet pre-training, it is currently limited by the relatively small scale of the pre-training dataset. As a direction for future research, we aim to explore cross-dataset pre-training to enlarge the pre-training dataset size and collaborate our framework with well-trained visual foundation models. These efforts are expected to overcome this limitation and enhance the generalizability and robustness of our framework.

{
\small
\bibliographystyle{ieeenat_fullname}
\bibliography{main}
}

\clearpage
\appendix
\section*{Appendix}
\section{Implementation Details}
\label{sec:implementation_details}

Our implementation is mainly based on Pointcept~\cite{pointcept}, a codebase focusing on 3D scene understanding and representation learning. The implementation details on pre-training and fine-tuning are listed below.

\subsection{Pre-training}
\label{sec:pretraining}

\vspace{1mm}
\mypara{Backbone architecture.}
Following previous self-supervised representation learning approaches~\cite{pointcontrast, csc, msc}, we adopt SparseUNet34C~\cite{spareseunet} as a backbone for ablation studies and result comparisons. The implementation detail of the backbone architecture is the same as in previous approaches.

\vspace{1mm}
\mypara{Pre-training dataset.}
Following previous work~\cite{pointcontrast, csc, msc}, we conduct self-supervised pre-training with GroupContrast on ScanNet v2~\cite{scannet} point cloud data.

\vspace{1mm}
\mypara{Data augmentation.}
We follow MSC~\cite{msc} to set our data augmentation pipeline for all experiments, which include Spatial augmentations, photometric augmentations and sampling augmentations. The data augmentation pipeline is illustrated in \tabref{tab:sup_aug}. 

\vspace{1mm}
\mypara{Pre-training setting.}
For ablation studies experiments, the number of default pre-training epochs is 600. For transfer learning results comparison, the number of pre-training epochs is 1200. Please refer to \tabref{tab:sup_pretrain} for more implementation details at the pre-training stage.

\begin{table}[h!]
\centering
\tablestyle{4pt}{1.1}
\begin{tabular}{l|cc}\toprule
Augmentation &Value \\\midrule
random rotate &angle=[-1, 1], axis=`z', p=1 \\
random rotate &angle=[-1/64, 1/64], axis=`x', p=1 \\
random rotate &angle=[-1/64, 1/64], axis=`y', p=1 \\
random flip &p=0.5 \\
random coord jitter &sigma=0.005, clip=0.02 \\
random color brightness jitter &ratio=0.4, p=0.8 \\
random color contrast jitter &ratio=0.4, p=0.8 \\
random color saturation jitter &ratio=0.2, p=0.8 \\
random color hue jitter &ratio=0.02, p=0.8 \\
random color gaussian jitter &std=0.05, p=0.95 \\
voxelization &voxel size=0.02 \\
random crop &ratio=0.6 \\
\bottomrule
\end{tabular}

\caption{\textbf{Data augmentation pipeline.}}
\label{tab:sup_aug}
\end{table}

\begin{table*}[t]
    \centering
    \subfloat[
        \textbf{Self-supervised pre-training on ScanNet}
        \label{tab:sup_pretrain}
    ]{
        \begin{minipage}{0.3\linewidth}{\begin{center}
            \tablestyle{4pt}{1.05}
            \begin{tabular}{l|c}
\toprule
Config &Value \\\midrule
optimizer &SGD \\
scheduler &cosine \\
learning rate &0.1 \\
weight decay &1e-4 \\
optimizer momentum &0.8 \\
batch size &32 \\
warmup epochs &12 \\
epochs &1200 \\
\bottomrule
\end{tabular}

        \end{center}}\end{minipage}
    }
    \hspace{3mm}
    \subfloat[
        \textbf{Semantic Segmentation fine-tuning on ScanNet}
        \label{tab:sup_semseg_scannet}
    ]{
        \begin{minipage}{0.3\linewidth}{\begin{center}
            \tablestyle{4pt}{1.05}
            \begin{tabular}{l|c}
\toprule
Config  &Value \\\midrule
optimizer &SGD \\
scheduler &cosine \\
learning rate &0.05 \\
weight decay &1e-4 \\
optimizer momentum &0.9 \\
batch size &48 \\
warmup epochs &40 \\
epochs &800 \\
\bottomrule
\end{tabular}

        \end{center}}\end{minipage}
    }
    \hspace{3mm}
    \subfloat[
        \textbf{Semantic Segmentation fine-tuning on S3DIS}
        \label{tab:sup_semseg_s3dis}
    ]{
        \begin{minipage}{0.3\linewidth}{\begin{center}
            \tablestyle{4pt}{1.05}
            \begin{tabular}{l|c}
\toprule
Config  &Value \\\midrule
optimizer &SGD \\
scheduler &cosine \\
learning rate &0.1 \\
weight decay &1e-4 \\
optimizer momentum &0.9 \\
batch size &12 \\
warmup epochs &0 \\
epochs &3000 \\
\bottomrule
\end{tabular}

        \end{center}}\end{minipage}
    }
    \\
    \centering
    \subfloat[
        \textbf{Instance Segmentation fine-tuning on ScanNet}
        \label{tab:sup_insseg_scannet}
    ]{
        \begin{minipage}{0.3\linewidth}{\begin{center}
            \tablestyle{4pt}{1.05}
            \begin{tabular}{l|c}
\toprule
Config  &Value \\\midrule
optimizer &SGD \\
scheduler &poly \\
learning rate &0.1 \\
weight decay &1e-4 \\
optimizer momentum &0.9 \\
batch size &48 \\
warmup epochs &0 \\
epochs &800 \\
\bottomrule
\end{tabular}

        \end{center}}\end{minipage}
    }
    \hspace{3mm}
    \subfloat[
        \textbf{Instance Segmentation fine-tuning on S3DIS}
        \label{tab:sup_insseg_s3dis}
    ]{
        \begin{minipage}{0.3\linewidth}{\begin{center}
            \tablestyle{4pt}{1.05}
            \begin{tabular}{l|c}
\toprule
Config  &Value \\\midrule
optimizer &SGD \\
scheduler &poly \\
learning rate &0.1 \\
weight decay &1e-4 \\
optimizer momentum &0.9 \\
batch size &12 \\
warmup epochs &0 \\
epochs &3000 \\
\bottomrule
\end{tabular}

        \end{center}}\end{minipage}
    }
    \hspace{3mm}
    \subfloat[
        \textbf{Object Detection fine-tuning on ScanNet and SUN-RGBD}
        \label{tab:sup_det_scannet}
    ]{
        \begin{minipage}{0.3\linewidth}{\begin{center}
            \tablestyle{4pt}{1.05}
            \begin{tabular}{l|c}
\toprule
Config  &Value \\\midrule
optimizer &SGD \\
scheduler &step \\
learning rate &1e-3 \\
weight decay &0 \\
optimizer momentum &0.9 \\
batch size &64 \\
warmup epochs &0 \\
epochs &180 \\
\bottomrule
\end{tabular}

        \end{center}}\end{minipage}
    }
    \caption{\textbf{Experiment settings.} We list experiment settings for both upstream pre-training and downstream fine-tuning.}
    \label{tab:ablation}
\end{table*}

\subsection{Fine-tuning}
\label{sec:finetune}

\vspace{1mm}
\mypara{Semantic segmentation.}
We use a SparseUNet~\cite{spareseunet} together with a projection layer for semantic segmentation fine-tuning. 
Experiments are conducted on ScanNet v2 and S3DIS. For ScanNet v2, we fine-tune the model on the training set and report the performance on the validation set. For S3DIS, we report the performance on Area 5 and use other data for fine-tuning.
For ScanNet and ScanNet200 semantic segmentation, the model is fine-tuned for 800 epochs with a batch size of 48. For S3DIS semantic segmentation, the model is fine-tuned for 3000 epochs with a batch size of 12. The voxel size is set to 0.02 for ScanNet fine-tuning and 0.05 for S3DIS fine-tuning. Please refer to \tabref{tab:sup_semseg_scannet} and \tabref{tab:sup_semseg_s3dis} for more details on semantic segmentation fine-tuning.
For data-efficient semantic segmentation on ScanNet, we follow the same setting as full dataset fine-tuning, as illustrated in \tabref{tab:sup_semseg_scannet}.

\vspace{1mm}
\mypara{Instance segmentation.}
We use SparseUNet~\cite{spareseunet} as the backbone and PointGroup~\cite{pointgroup} as the segmentation head for instance segmentation fine-tuning. 
Experiments are conducted on ScanNet v2 and S3DIS. For ScanNet v2, we fine-tune the model on the training set and report the performance on the validation set. For S3DIS, we report the performance on Area 5 and use other data for fine-tuning.
For ScanNet and ScanNet200 instance segmentation, the model is fine-tuned for 800 epochs with a batch size of 48. For S3DIS instance segmentation, the model is fine-tuned for 3000 epochs with a batch size of 12. The voxel size is set to 0.02 for ScanNet fine-tuning and 0.05 for S3DIS fine-tuning. Please refer to \tabref{tab:sup_insseg_scannet} and \tabref{tab:sup_insseg_s3dis} for more details on instance segmentation fine-tuning.

\vspace{1mm}
\mypara{Object detection.}
We use SparseUNet~\cite{spareseunet} as the backbone and VoteNet~\cite{votenet} as the detection head for object detection fine-tuning. 
Experiments are conducted on ScanNet v2 and SUN-RGBD. We fine-tune the model on the training set and report the performance on the validation set.
We report the transfer learning results on ScanNet and SUN-RGBD object detection. We fine-tune the model for 360 epochs with a batch size of 64 for both datasets. The voxel size is set to 0.02. Please refer to \tabref{tab:sup_det_scannet} for more details on object detection fine-tuning.

\begin{figure*}[t]
\centering
\includegraphics[width=0.95\linewidth]{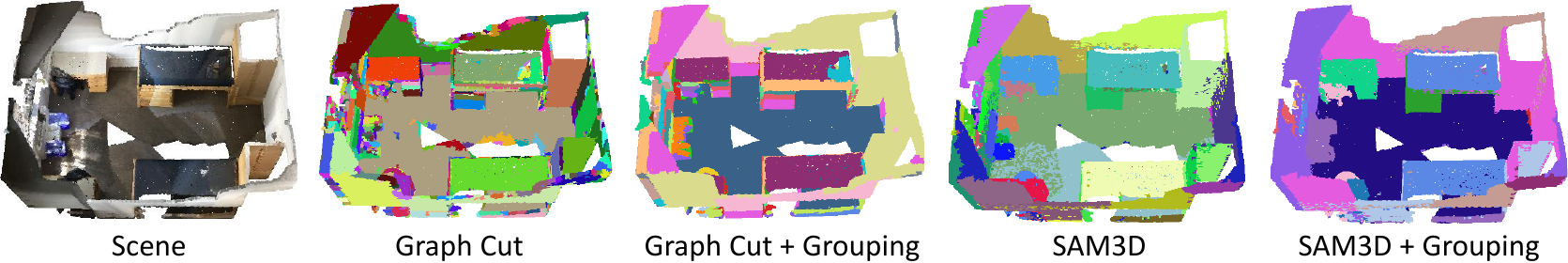}
\caption{Segment Grouping is capable of aggregating both Graph Cut mask and SAM3D mask into semantic meaningful regions.}
\label{fig:sam_gc}
\end{figure*}

\section{Collaboration with Foundation Models}
\label{sec:failure_cases}
We further study the potential of collaborating our work with existing visual foundation models, such as Segment Anything Models (SAM)~\cite{sam}.
Recently, there emerge several works that leverage SAM to predict 3D bounding boxes or segmentation masks on point clouds. These segmentation masks can directly replace the GraphCut~\cite{graph_segment} results in Segment Grouping.
To assess this possibility, we substitute the GraphCut results with the segmentation mask of SAM3D~\cite{yang2023sam3d} and validate its effectiveness on 3D representation learning.
As depicted in \figref{fig:sam_gc}, Segment Grouping successfully clusters both Graph Cut mask and SAM3D mask into proper regions.
The mIoU result for ScanNet-v2 semantic segmentation fine-tuning is $\mathbf{75.9\%}$, which is higher than the result that using Graph Cut ($\mathbf{75.7\%}$). 
Incorporating existing visual foundation models is a promising way to mitigate data scarcity for 3D visual representation learning. We intend to pursue further in our future research.

\section{Prototype Visualization and Analysis}
\label{sec:proto_vis}
\begin{figure*}[t]\centering
\includegraphics[width=0.95\linewidth]{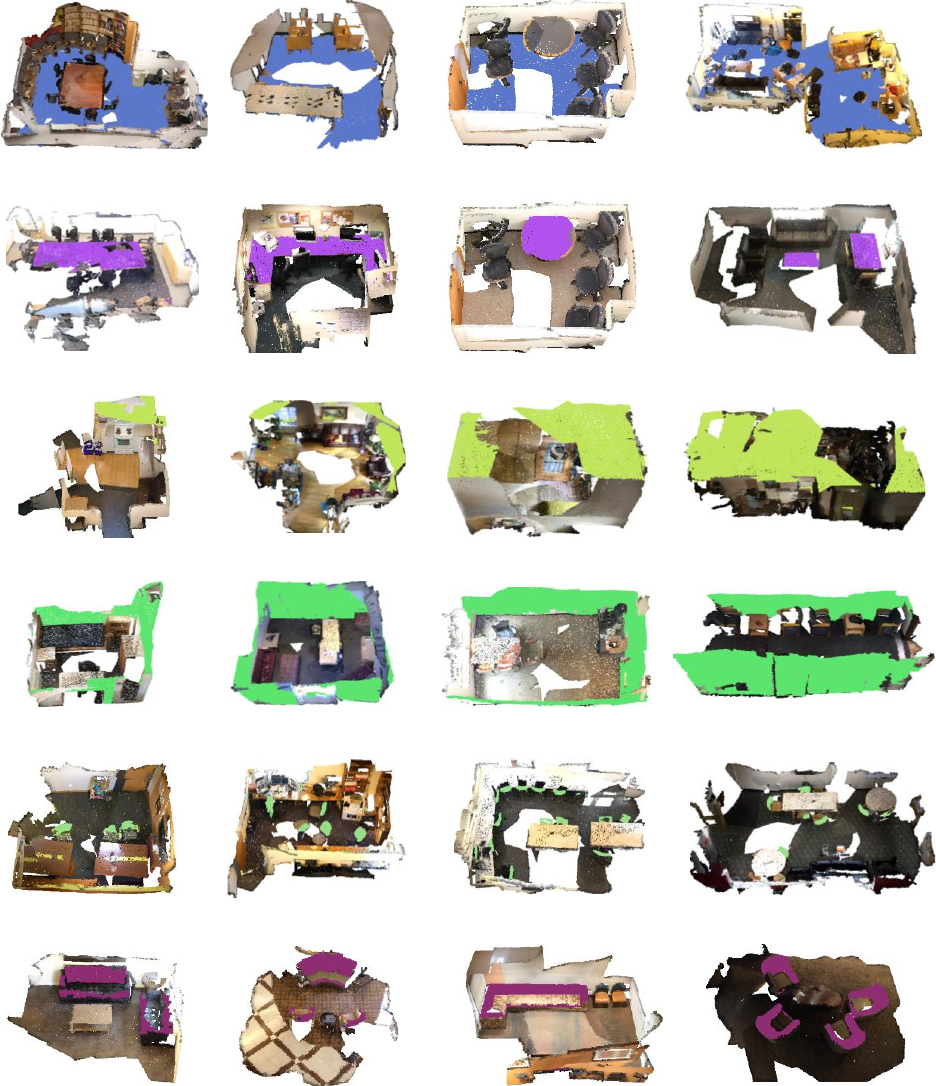}
\caption{\textbf{Prototype Visualization.} Each row refers to one prototype, and the group regions are highlighted with a specific color. Our method can discover semantic meaningful concepts from unlabeled 3D scenes. }
\label{fig:prototype}
\end{figure*}

We attempt to visualize the regions assigned to each prototype to analyse whether the randomly initialized prototypes can learn semantic meanings. As illustrated in \figref{fig:prototype}, the model successfully discovers semantic meaningful concepts from unlabeled 3D scenes. These concepts include semantic categories such as floor, table, ceiling and wall, as well as object parts like chair backrests and sofa backrests. The visualization results demonstrate that the prototypes have effectively learned and captured semantic meaning.

Since no supervision signals are provided, the results of segment grouping are bound to orthogonal to the semantic labels sometimes. 
For example, assign points with the same semantic label to different clusters, or group points with different semantic labels into identical clusters.
We believe discovering semantic meaningful subcategories is not harmful at the representation learning stage. It can help the model learn a better representation space and benefit downstream fine-tuning.

\end{document}